%% file: fedgrains_arxiv.tex
\documentclass[twoside,leqno,twocolumn]{article}

\usepackage[letterpaper]{geometry}
\usepackage[sort]{cite}
\usepackage{ltexpprt}
\usepackage{hyperref}
\usepackage{amsmath}
\usepackage{cuted}
\usepackage[dvipsnames]{xcolor}
\usepackage{amsfonts, bm}
\usepackage{booktabs}
\usepackage{graphicx}
\usepackage{amsfonts}
\usepackage{algorithm}
\usepackage{algorithmicx}
\usepackage{algpseudocode}
\usepackage{appendix}
\usepackage{minitoc}

\begin{document}

\newcommand\relatedversion{}
\renewcommand\relatedversion{\thanks{The full version of the paper can be accessed at \protect\url{https://arxiv.org/abs/1902.09310}}} 

\title{\Large FedGrAINS: Personalized SubGraph Federated Learning \\with AdaptIve Neighbor Sampling}

\author{Emir Ceyani \\ University of Southern California 
\and Han Xie  \\ Emory University \and Baturalp Buyukates \\ University of Birmingham 
\and Carl Yang \\Emory University
\and Salman Avestimehr\\ University of Southern California }

\date{}

\maketitle

\fancyfoot[R]{\scriptsize{Copyright \textcopyright\ 2025 by SIAM\\
Unauthorized reproduction of this article is prohibited}}





\input{sections/abstract}

\input{sections/intro.tex}
\input{sections/problem.tex}
\input{sections/fedgrains.tex}

\input{sections/exps.tex}
\input{sections/related.tex}
\input{sections/conclusion.tex}

\bibliographystyle{siam}
\bibliography{references.bib}

\clearpage
\appendix
\onecolumn
\input{sections/appendix.tex}

\end{document}

%% file: sections/abstract.tex
\begin{abstract}
Graphs are crucial for modeling relational and biological data. As datasets grow larger in real-world scenarios, the risk of exposing sensitive information increases, making privacy-preserving training methods like federated learning (FL) essential to ensure data security and compliance with privacy regulations. Recently proposed personalized subgraph FL methods have become the de-facto standard for training personalized Graph Neural Networks (GNNs) in a federated manner while dealing with the missing links across clients' subgraphs due to privacy restrictions. However, personalized subgraph FL faces significant challenges due to the heterogeneity in client subgraphs, such as degree distributions among the nodes, which complicate federated training of graph models. To address these challenges, we propose \textit{FedGrAINS}, a novel data-adaptive and sampling-based regularization method for subgraph FL. FedGrAINS leverages generative flow networks (GFlowNets) to evaluate node importance concerning clients' tasks, dynamically adjusting the message-passing step in clients' GNNs. This adaptation reflects task-optimized sampling aligned with a trajectory balance objective. Experimental results demonstrate that the inclusion of \textit{FedGrAINS} as a regularizer consistently improves the FL performance compared to baselines that do not leverage such regularization. 
\end{abstract}

\subsection*{Keywords:} federated learning, graph neural networks, reinforcement learning, node classification, generative models, generative flow networks

%% file: sections/intro.tex
\section{Introduction}
\label{intro} 

Graph Neural Networks (GNNs) operate over a single graph, with nodes and edges stored in a central server. For example, in social networks, each user, along with 
their connections form a large graph of all users and their links \cite{alam2023fedaiot}. However, parties may not send their private graph datasets to a central server due to privacy concerns, raising the need to train GNN models over multiple distributed graph datasets. For instance, hospitals may maintain their patient interaction networks to track physical contacts or co-diagnoses of diseases. However, they cannot share these graphs with other entities due to privacy restrictions~\cite{FedSage}. Here, the question is how to collaboratively train GNNs across clients' distributed subgraphs without sharing the actual graph data. Federated Learning (FL) of GNNs aims to solve this problem so that each client individually trains a local GNN on their local data, while a central server aggregates locally updated weights from multiple clients into a global model using federated optimization \cite{he2021fedgraphnn}. 

A key challenge in na\"{i}vely applying FL methods to GNNs is the potential loss of crucial information due to missing links between subgraphs distributed across parties. Recent approaches to subgraph FL address this issue by extending local subgraphs with information from other subgraphs \cite{FedSage,FedGNN}. Specifically, they extend the local subgraph by either precisely adding the relevant nodes from other subgraphs from different clients \cite{FedGNN} or by predicting the nodes using node information from the other subgraphs via graph mending \cite{FedSage}. However, sharing node information can compromise data privacy and lead to high communication costs. While \cite{FEDPUB} mitigates the issues of missing links and subgraph heterogeneity through personalized aggregation and local sparse masks, all existing approaches are mainly based on image data modality, failing to fully exploit the graph-structured data to address heterogeneity, thus resulting in sub-optimal solutions to personalize GNN models.  

On the other hand, dropout-based data pruning in FL\cite{dropout,FJORD} and stochastic regularization in GNNs\cite{li2023less,hasanzadeh2020bayesian} have shown to be successful in regularizing deeper models without any significant overhead by randomly dropping the network weights and/or data components. Existing dropout-based data pruning approaches in FL, however, are not suitable for graph ML models due to GNNs' invariance requirements, which forces us to utilize dropout methods for graph-structured data \cite{Rong2020DropEdge,fang2023dropmessage}. Further, applying dropout techniques na\"{i}vely to deeper GNNs without accounting for the data structure can degrade node and structural information quality due to Laplacian smoothing, leading to over-smoothened representations that are uninformative in higher layers\cite{li2018deeper}. Moreover, using the same dropout parameters across different clients is also not an effective strategy due to heterogeneity over data, system, and the model, raising the need for coordination based on the users' requirements. 

To effectively personalize GNN models by respecting graph structures, we propose a novel data-adaptive, sampling-based regularization method for subgraph FL, named \textit{FedGrAINS} by estimating and selecting essential nodes at each layer of GNNs every round for every client in a federated manner. To estimate the adaptive neighborhood distribution of each node in the datasets, we leverage Generative Flow Networks (GFlowNets), the state-of-the-art generative models tailored for structured discrete objects like graphs \cite{bengio2021gflow-fundation}. Here, for each node in the clients' graph, we leverage GFlowNets to estimate the importance of the neighbors of a node so that we can prune the graph for each node adaptively in parallel to optimize the clients' personalized task performances. We refer to this framework as Personalized Sub\textit{Gr}aph Federated Learning with \textit{A}dapt\textit{I}ve \textit{N}eighbor \textit{S}ampling (\textit{FedGrAINS}). We extensively validate the performance of FedGrAINS across six popular graph datasets, evaluating overlapping and disjoint subgraph FL scenarios with varying numbers of clients. Our comprehensive experimental results show that FedGrAINS significantly outperforms relevant baselines. Our contributions are summarized as follows:
\begin{itemize}
    \item  We introduce a new type of heterogeneity, \textit{node-degree heterogeneity}, showing that it cannot be handled with previous personalized FL approaches.
    \item  We propose \textit{FedGrAINS}, a novel framework for personalized subgraph FL. \textit{FedGrAINS} performs layer-wise node-importance estimation using GFlowNets and dynamically updates the GNN message-passing steps to combat neighborhood heterogeneity caused by missing links and subgraph heterogeneity between multiple clients.
    \item  We validate our personalized FL framework \textit{FedGrAINS} across six real-world datasets under disjoint and overlapping node scenarios, demonstrating its effectiveness against relevant baselines.
\end{itemize}

%% file: sections/problem.tex
\section{Preliminaries}
\label{sec:formulation}
In this section, we describe GNNs and personalized subgraph FL in more detail. Then, we present the core component of our proposed algorithm, GFlowNets.

\subsection*{GNNs:}
Let $\mathcal{G}=(\mathcal{V,E})$ be an undirected graph with a set of $N$ nodes $\mathcal{V}=\left\{v_{1}, \dots v_{N}\right\}$ and edges $\mathcal{E}$. $\bm{X} = \left\{\bm{x}_{1}, \dots ,  \bm{x}_{N}\right\} \in \mathbb{R}^{N \times d}$, where $\mathbf{x}_{i} \in \mathbb{R}^{d}$ is the node feature vector for node $v_{i}$, indicating the features of each node $i \in 
\{1,\ldots, N \}$. According to the Message-Passing Neural Network (MPNN) paradigm \cite{MPNN}, the flow of hidden representations in GNNs, based on the nodes' neighborhoods and features, is described as follows:
\begin{equation}
\fontsize{9pt}{9pt}\selectfont
    \bm{H}^{l+1}_v
    = \text{UPD}^{l}\left( \bm{H}^{l}_v, \text{AGG}^{l}\left(\left\{  \bm{H}^{l}_u: \forall u\in\mathcal{N}(v) \right\}\right) \right),
    \label{eq:mp}
\fontsize{9pt}{9pt}\selectfont
\end{equation}
where $\bm{H}^{l}_v$ denotes the features of node $v$ at the $l$-th layer, and $\mathcal{N}(v)$ denotes the set of neighbor nodes of node $v$: $\mathcal{N}(v) = \left\{ u \in \mathcal{V} \; | \; (u, v) \in \mathcal{E} \right\}$. $\text{AGG}$ aggregates the features of the neighbors of node $v$. We note that  $\text{AGG}$ can be mean or even a black-box function. $\text{UPD}$ updates node $v$'s representation given its previous representation and the aggregated representations from its neighbors. $\bm{H}^{1}$ is initialized as $\bm{X}$.

As the number of layers increases, the embedding computation for a node $v$ incorporates information from neighbors several hops away.
Increasing the number of layers expands the depth of neighbors for each node included in the training set, resulting in the exponential increase in the neighborhood size for each node, thus causing high computation time on the user side. Reducing high computation at the edge is possible by training GNNs in a subgraph federated learning setting where each user has a part of an entire graph and utilizes FL to access higher-order node information. The following subsection describes the federated training process for GNNs.

\subsection*{Personalized Subgraph FL:} We assume that there exists a global graph $\mathcal{G}_{\textit{global}}=(\mathcal{V}_{global},\mathcal{E}_{global})$, where $\mathcal{V}_{global}$ and $\mathcal{E}_{global}$
are the set of nodes and edges of the global graph, respectively. $\bm{X}_{global}$ is the associated node feature set of the global graph. In the subgraph FL system, we
have one central server $S$ and $M$ clients with distributed
subgraph datasets: $\mathcal{D}_i = \{\mathcal{G}_i, \bm{Y}_i \} = \{\{\mathcal{V}_i, \mathcal{E}_i, \bm{X}_{i}\} , \bm{Y}_{i}\}$ where $\bm{X}_{i}$ and $\bm{Y}_{i}$ are node feature and label sets of the $i$-th client, respectively, for $i \in [M]$, where $\mathcal{V}_{global}=\bigcup_{i=1}^{M} \mathcal{V}_i$. For
an edge $e_{v,u} \in \mathcal{E}_{global} $, where $v \in \mathcal{V}_i , u \in \mathcal{V}_j$, we have
$e_{v, u} \notin \mathcal{E}_i \cup \mathcal{E}_j$. That is, $e_{v, u}$ might exist in reality but
is missing from the whole system, as this link is between the local data of two distinct clients $i$ and $j$. The system exploits
an FL framework to collaboratively learn $M$ personalized node
classifiers $f_i$, parameterized with $\phi_i$, on isolated subgraphs in clients without raw
graph data sharing. Finding personalized models is equivalent to minimizing the aggregated empirical risk defined as:
\begin{equation}
    \min _{\phi_{1},\dots\phi_{M}} \frac{1}{M} \sum_{i=1}^M \mathcal{R}_i\left(f_i(\phi_{i})\right),
\end{equation} where the empirical risk for the $i$-th client is defined as  $\mathcal{R}_i\left(f_i(\phi_{i})\right) = \mathbb{E}_{\left(\mathcal{G}_i, \bm{Y}_i\right) \sim \mathcal{D}_{\text {i}}}\left[\mathcal{L}_i\left(f_i\left(\phi_{i} ; \mathcal{G}_i\right), \bm{Y}_i\right)\right]$, and $\mathcal{L}_i\left(f_i\left(\phi_{i} ; \mathcal{G}_i\right), \bm{Y}_i\right)$ is the $i$-th client's average loss over the training data.

\subsection*{GFlowNets:}

GFlowNets are the state-of-the-art generative models for generating structured objects, such as molecules and graphs, from a finite set of states $\mathcal{S}$§
in proportion to a reward function $R(s_{T}): \mathcal{S}_{T}\rightarrow \mathbb{R}_+$, that is assigned to terminal states $s_{T}$ \cite{bengio2021gflow-fundation}. At a particular time $t$, based on the current state $s_t$, we perform an action $a_t$ to propagate into the next state $s_{t+1}$ with the help of the forward transition function $P_{F}$. The consecutive state-action pairs starting from an initial state $s_0 \in \mathcal{S}_0 \subset \mathcal{S}$ and ending at a terminal state $s_{T} \in \mathcal{S}_{T} \subset \mathcal{S}$ form a trajectory $\tau = (s_{0}, a_{0}, \dots , s_{T})$.  The key idea in GFlowNets is to learn a flow network function $F(s)$ that ensures that the total flow into each state equals the flow out. Hence, the probability of reaching any terminal state is proportional to its
reward $R(s_n)$, achieved by aligning the forward and backward policies $P_{F}$ and $P_{B}$, parameterized with neural networks, ensuring that the
final distribution over terminal states reflects the desired reward distribution. Therefore, the GFlowNet learning problem is equivalent to learning forward 
and backward policies $\pi  = \{ P_F(s_t|s_{t-1}), P_B(s_{t-1}|s_t)\}$. Trajectory balance (TB) \cite{malkin2022trajectorybalance} is the most common learning objective designed to
improve credit assignment and learning policies in GFlowNets. Given $\tau$, the TB is defined as: 
\begin{equation}
\label{trajectory balance}
    \mathcal{L}_{TB}(\tau) = \left (\log \frac{Z(s_0) \prod_{t=1}^n P_F(s_t|s_{t-1})}{R(s_n)\prod_{t=1}^n P_B(s_{t-1}|s_t)} \right)^2,
\end{equation}
where $Z(s_o):\mathcal{S}_0\rightarrow \mathbb{R}_+$ computes the total flow of the network from starting state $s_0$, which is also known as the normalizing constant. 

In the next section, we outline our GFlowNet modeling approach to characterize the neighborhood importance distribution for each node and explain how we learn and balance global and local neighborhood distributions in federated settings.

%% file: sections/fedgrains.tex
\section{Personalized Subgraph Federated Learning with Adaptive Neighbor Sampling}

\begin{figure*}[t]
    \centering
    \includegraphics[width=0.90\linewidth]{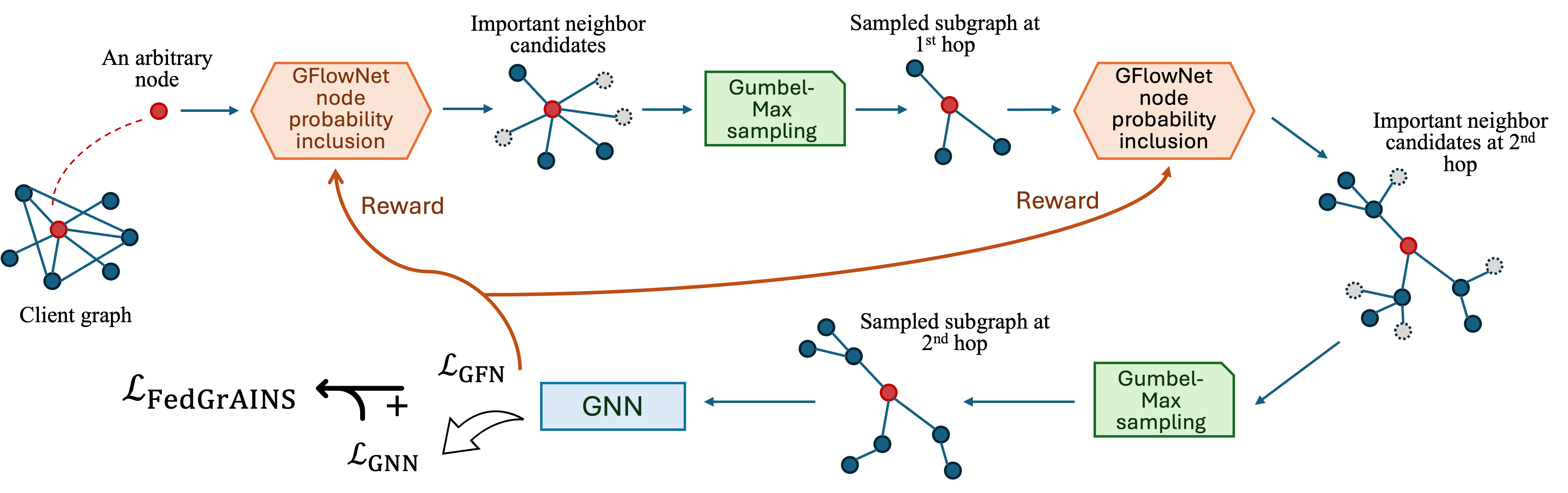}
    \caption{The illustration of the proposed joint training scheme of 2-layer GNN and GFlowNet for our personalized subgraph FL framework, \textit{FedGrAINS} from the client side. For any node $v$ in the client's graph, we first estimate the important neighbors with GFlowNet, then sample $k$ nodes using the Gumbel-Max trick \cite{huijben2022review-gumbel}. After choosing the important nodes for two layers, we input the sampled subgraph to the GNN. After obtaining our novel loss, we backpropagate through the classification and the trajectory balance losses, with the reward function defined as the cross-entropy loss at a specific round. With classification loss as a reward function, GFlowNet adapts itself to select and sample neighbors vital for optimal task performance.  }\label{fig:fedgrains sketch}
\end{figure*}

In this section, we introduce \textit{FedGrAINS}, our personalized subgraph FL method. We detail the design of GFlowNet to estimate neighborhood importance distributions for each node (Section \ref{sec:gflownet-design}), the scalable sampling procedure (Section \ref{sampling and off policy}), and the personalized subgraph FL framework \textit{FedGrAINS} (Section \ref{fedgrains}).
 
\subsection{Adaptive Neighborhood Distribution Estimation with GFlowNets:} \label{sec:gflownet-design}

Given graph data, the adjacency matrix $\bm{A} \in \mathbb{R}^{N \times N}$ allows us to determine the neighbors of every node $v \in \mathcal{V}$. At a given layer $\ell$, the $\text{AGG}$ operator given in (\ref{eq:mp}) aggregates the features of $v$'s neighbors based on $\bm{A}$ and requires higher-order propagation to compute neighbors' messages \cite{GNN}. As a result, the neighborhood size grows exponentially with the number of layers, resulting in over-smoothing.
It is possible to regularize models based on randomly dropping nodes, edges, and even messages\cite{Rong2020DropEdge,fang2023dropmessage}. However, these methods operate independently of the task, data, and classifier, leading to suboptimal performance \cite{li2018deeper}. Guided by the intuition that the aggregation step in message-passing GNNs is crucial for good performance, we aim to adaptively choose and sample the neighbors at every layer to minimize the classification loss using GFlowNets. Since every MPNN layer requires an adjacency matrix to compute hidden representations, we sample a sequence of adjacency matrices over all layers. Thus, for GFlowNet to sample important neighbors, we define the state $s \in \mathcal{S}$ as a sequence of adjacency matrices $s = (\bm{A}_0, \ldots, \bm{A}_{\ell})$ sampled up to the current step. The action $a_{\ell}$ to construct the adjacency matrix for the next layer $\bm{A}_{\ell + 1}$ is to choose $k$ nodes without replacement among the neighbors of nodes in $\bm{A}_{\ell}$. Then, in our design, obtaining the adjacency matrix of the next layer is equivalent to transitioning to a new state.  

Using the modeling described above, in a $L$-layer GNN, we construct a sequence of $L$ adjacency matrices to reach a terminating state. We aim to perform task-aware sampling with the optimal GFlowNet sampling policy to minimize the expected classification loss  : \vspace{-1mm}
\begin{align}
    R(s_L) &= R(\bm{A}_0, \ldots, \bm{A}_L)  \nonumber\\
    &= \exp(-\alpha\cdot \mathcal{L}\left(F\left(\phi ; \mathcal{G}\right), \bm{Y}\right)) \nonumber\\
    &= \exp(-\alpha\cdot \mathcal{L}\left(F\left(\phi ; \mathcal{V} , (\bm{A}_{0}, \ldots, \bm{A}_{L})\right), \bm{Y}\right)), 
\end{align} where $\alpha$ is a \emph{scaling parameter}. $\bm{A}_{0}$ is the adjacency matrix of the graph at the starting state, and $\bm{A}_{1},\ldots, \bm{A}_{L}$ are the adjacency matrices sampled at every layer. 

\vspace{-4mm}
\subsubsection*{Forward Probability:}
The GFlowNet of \textit{FedGrAINS} is a neural network that learns the forward probabilities $P_F(s_{\ell+1}|s_{\ell})$, the probability of sampling an adjacency matrix $\bm{A}_{\ell +1}$ for the layer $\ell +1$ given the previous layers' adjacency matrices $\bm{A}_{0}, \dots ,\bm{A}_{\ell}$ . For each node $v_i$, where $i \in {1, \ldots, N}$, the GFlowNet models the probability that node $v_i$ in the neighborhood of the sampled nodes at layer $\ell$, $\mathcal{N}(\mathcal{S}^{\ell})$, will be included in $\mathcal{V}^{\ell+1}$, the set of nodes sampled for layer $\ell+1$. We represent the inclusion probability of each node in the sample using a Bernoulli distribution \cite{fang2023dropmessage, Rong2020DropEdge}. 
In this formulation, $P_F(s_{l+1}|s_l)$  is tractable thanks to Bernoulli parameterization for each node associated with a logit indicating its likelihood of being included in the sampled set. Thus, if $p_i$ denotes the probability that node $v_i$ in the neighborhood of the sampled nodes at layer $\ell$ will be included in $\mathcal{V}^{\ell +1}$, then the forward probability transition function between layers $\ell$ and $\ell+1$ is
$P_F(s_{\ell+1}|s_\ell)=\prod_{v_i\in \mathcal{V}^{\ell+1}} p_{i}\prod_{v_j \in \mathcal{N}(\mathcal{S}^{\ell}) \backslash \mathcal{V}^{\ell +1}} (1-p_j)$, where the first product computes the probabilities of the nodes that are in the newly sampled set $\mathcal{V}^{\ell+1}$, and the second one computes the probabilities of nodes that are not within the newly sampled set $\mathcal{V}^{\ell+1}$.
\vspace{-4mm}
\subsubsection*{Backward Probability:}
TB requires us to define the backward transition function $P_B(s_{\ell}|s_{\ell+1})$ \cite{malkin2022trajectorybalance}. 
However, in our setup, it is not required as the state representation, $s=(\bm{A}_{0}, \dots, \bm{A}_{\ell})$, exactly saves the trajectory that was taken through the flow network to get to the state $s$. Thus, the backward mapping is unique and deterministic. So, $P_B(s_{l}|s_{l+1})=1$.
\vspace{-4mm}
\subsubsection*{Trajectory Balance Condition:} Given $\tau$ and model weights $\phi_{GFN}$, we define TB condition \cite{malkin2022trajectorybalance} as:
\scalebox{0.4}{}{%
\begin{align} 
&\mathcal{L}^{\textit{GFN\!}}(\tau ; \phi_{GFN\!}) \!=\! \left (\log \frac{Z\prod_{l=1}^L P_F(s_l|s_{l\!-\!1} ; \phi_{GFN\!})}{R(s_L)} \right)^{\!2} \nonumber\\
&\!=\! \left( \log Z \!+\! \sum_{l=1}^L \log P_F(s_l|s_{l\!-\!1} ; \phi_{GFN\!}) \!+\! \alpha \!\cdot\! \mathcal{L}^{GNN\!} \right)^{\!2}\!\!. \label{eq:trajectoryshort}
\end{align}
}Notice that we model the normalizer $Z(s_0)$ in (\ref{trajectory balance}) as a constant in our loss formulation to skip the additional optimization during parameter estimation. After estimation, we describe the node sampling procedure of our proposed personalized subgraph FL framework.
 
\subsubsection{Sampling and Off-Policy Training:}
\label{sampling and off policy}
We want to sample $k$ important ones from $N$ graph nodes without replacement at every message passing step. However, our forward transition function $P_F$ contains independent Bernoulli distributions $N$, and it is unlikely that we will sample exactly $k$ nodes of this distribution. 
Instead, we use the Gumbel-max trick \cite{xie2019reparameterizable,kool2019stochastic, huijben2022review-gumbel}, which selects a set of nodes $\mathcal{V}_k^{\ell}$ by perturbing the log probabilities randomly and taking the top-$k$ among those: 
\begin{align}
    \mathcal{V}_k^{l} = \mathsf{top}(k, \log p_1+G_1, ..., \log p_{n} + G_{n}), 
    \label{eqn:gumbel}
\end{align} 
where $G_{i}\sim\text{Gumbel}(0,1)$. The sampling process shown above differs from the sequential sampling of the forward transition function $P_F$ one node at a time. Thus, we train the GFlowNet in an off-policy setting using the trajectory balance loss without any change \cite{malkin2022gflownets-off}. In addition, the off-policy training is less expensive in computation and more robust to high-variance problems in distribution estimation than direct gradient-estimation methods such as REINFORCE\cite{ahmed2023simple}. Next, we explain how the adaptive importance sampling framework assists the (personalized) subgraph FL in Section \ref{fedgrains}.

\subsection{Federated Training of Node Classifier and Importance Sampler:} 
\label{fedgrains} 

In \textit{FedGrAINS}, each client desires an optimal yet personalized performance in the node classification task while not sharing any data with others. For this, we formulate the joint training of the node classifier and the importance sampler as depicted in Figure \ref{fig:fedgrains sketch}. First, for each layer of local GNN, we estimate the important nodes to build layer-wise adjacency matrices for each layer using the GFlowNet and Gumbel-max trick as defined in Section \ref{sampling and off policy}. Then, using the estimated layer-wise adjacency matrices, we perform node classification using the local model. More details on the forward pass and the training loop of \textit{FedGrAINS} client can be found in Appendix \ref{appendix:fedgrains-algo}. For each client, we formulate the following personalized subgraph federated optimization: 
\begin{equation}
    \min_{\{\phi_{GNN}, \phi^{1}_{GFN}, \dots , \phi^{M}_{GFN} \} } \mathcal{L}^{i}_{\text{FedGrAINS}}\left(\phi_{GNN},\phi^{i}_{GFN} \right). \label{eqn:fedgrains}
\end{equation}  where $\mathcal{L}^{i}_{\text{FedGrAINS}}\left(\phi_{GNN},\phi^{i}_{GFN} \right)=\mathcal{L}^{GFN}_{i}+ \mathcal{L}^{GNN}$ is the loss function of the $i$-{th} client, $\mathcal{L}_{i}^{GFN}$ is defined as in Equation \ref{eq:trajectoryshort} and $\mathcal{L}^{GNN} = \mathcal{R}_i\left(f_i(\phi_{GNN})\right)$ is the classification loss.

%% file: sections/exps.tex
\section{Experiments}
\label{experiments}

\subsection{Experimental Setup}
\label{experiments:datasets}
\subsubsection*{Datasets} As proposed by \cite{FEDPUB, FedSage}, we generate distributed subgraphs by dividing the graph datasets into a certain number of clients so that each FL client has a subgraph that is a part of the original graph. Specifically, we use six datasets: Cora, CiteSeer, Pubmed, and ogbn-arxiv for citation graphs \cite{planetoid, ogb}; Computer and Photo for product graphs~\cite{amazon, amnazonsubset}. We partition the graphs into subgraphs using the METIS graph partitioning algorithm~\cite{Karypis95metis}, which allows specifying the number of subsets without requiring further merging, unlike Louvain partitioning \cite{blondel2008fast}. METIS first coarsens the original graph using maximal matching methods \cite{karypis1998fast}, then computes a minimum edge-cut partition on the coarsened graph followed by the projection of the partitioned coarsened graph back onto the original graph. In our experiments, we consider two subgraph settings:  \textit{Disjoint subgraphs} setting, in which each client has a unique set of nodes and no overlapping nodes between subgraphs. In this setup, we use the METIS outputs as they provide non-overlapping partitions. We consider 5, 10, or 20 clients for the disjoint setting. In the \textit{overlapping subgraphs} setting, clients may have common nodes between their datasets. Since METIS produces non-overlapping partitions, we introduce overlap by randomly sampling subgraphs multiple times from the partitioned graph. First, we divide the original graph into 2, 6, or 10 disjoint subgraphs using METIS. Then, for each METIS partition, we randomly sample half of the nodes and their associated edges five times to create subgraphs. This process results in 10, 30, or 50 clients for the overlapping case, generating shared nodes across subgraphs from the same METIS partition. It is important to note that the number of clients in our proposed settings is not uniform due to the need for overlapping nodes. For each subgraph, except for the ogbn-arxiv dataset, train, validation, and test node random sampling ratios are 20\%, 40\%, and 40\%, respectively. Thus, for the ogbn-arxiv dataset, train, validation, and test node random sampling ratios are 5\%, 47.5\%, and 47.5\%, respectively, as the ogbn-arxiv dataset has a significantly higher number of nodes, as in Table \ref{tab:appendix:data}. 

\subsubsection*{Baselines} \textbf{1) Local}: A baseline that locally trains models without weight sharing. \textbf{2) FedAvg}~\cite{McMahan2017CommunicationEfficientLO}: The most popular and naive FL baseline. \textbf{3) FedPer}~\cite{arivazhagan2019federated}: A personalized FL baseline without sharing personalized layers adopted to subgraph data. \textbf{4) FedSage+}~\cite{FedSage}: A subgraph FL baseline that proposes the subgraph FL problem with missing links and introduces the first solution by training local models with a global fake neighborhood generator. \textbf{5) GCFL}~\cite{GCFL}: A graph FL baseline which formulates graph level FL as clustered FL~\cite{sattler2020clustered}, adopted for subgraph FL. \textbf{6) FED-PUB}\cite{FEDPUB}: State-of-the-art personalized subgraph FL framework, which includes similarity matching and weight masking. \textbf{7) \textit{FedGrAINS}}: Our personalized subgraph FL framework with adaptive neighborhood sampling using GFlowNets. Further details for the baselines can be found in Appendix \ref{appendix:sub:model}.

\subsubsection*{Implementation Details} For our experiments, we employ two layers of Graph Convolutional Networks (GCNs)~\cite{GCN} with ReLU activation as the base GNN for the flow networks $P_F$ and $P_B$, as well as for the global and local models. The important hyperparameters are a hidden layer dimensionality of 128 and a learning rate of 0.01. We set the sparsity and proximal terms for the FED-PUB framework to 0.001 \cite{FEDPUB}. \textit{FedGrAINS} requires three additional hyperparameters: the sampling budget $k$, the learning rate for the GFlowNet, $\beta_{GFN}$, and the reward scaling
parameter $\alpha$. We set the reward scaling to $10^5$, while $\beta_{GFN}$ is set to 0.001 for all experiments (more detailed analysis for GFlowNet hyperparameters can be found in Section \ref{sec:gfn-hparam}). We tune these parameters based on classification performance on the validation set. We utilize Adam optimizer~\cite{Adam} for all models as local optimizer and perform FL (with full participation) over 100 communication rounds with one local training epoch for the Cora, CiteSeer, and Pubmed datasets and 200 rounds with three local training epochs for the Computer, Photo, and ogbn-arxiv datasets, considering the dataset size. We measure the node classification accuracy on subgraphs on the client side and then average over all clients.

\subsection{Experimental Results}

\subsubsection*{Main Results} 
Tables \ref{tab:main:nonoverlap} and \ref{tab:main:overlap} show node classification results in disjoint and overlapping subgraph scenarios, respectively. The primary motivation behind the proposed \textit{FedGrAINS} is its applicability to existing subgraph FL algorithms. For a more fair comparison with Fed-PUB \cite{FEDPUB}, we combine our data-adaptive regularization framework with their personalized aggregation and sparse masking. Other baselines do not include personalized aggregation. In addition, we did not test how the inclusion of our proposed framework affects the performance of FedSage+ as our proposed framework tries to filter unimportant nodes. In contrast, FedSage+ tries to increase the number of neighbors with fake neighbors, providing a more complicated setup for us to observe the effect of only learnable node filtering. Based on the experimental results, we make the following observations: First,
regardless of the FL algorithm, classification performance
decreases when the number of clients increases due to the increased number of partitions creating more missing links between the clients (See Appendix \ref{appendix:sub:data:links} for the missing links counts for each dataset). The increasing number of missing links also leads to severe node-degree heterogeneity(see Appendix \ref{appendix:sub:data:degree-het} )within clients. This results in more challenging local training and collaborative learning of generalizable models with other clients. 

Second, the disjoint setting is more challenging than the overlapping one for the same number
of clients because the subgraphs in the non-overlapping setting are entirely disjoint and more heterogeneous than the overlapped case. Moreover, the non-overlapping setting has fewer nodes to learn from due to the overlap generation scheme explained in the experimental design. As shown in Table~\ref{tab:main:nonoverlap}, \textit{FedGrAINS} consistently outperforms all existing baselines in this challenging scenario, demonstrating its efficacy. 

Finally, the inclusion of \textit{FedGrAINS} as a personalization method consistently improves the performance of all baselines. Specifically, the GFlowNet-based importance sampler improves the accuracy of all algorithms (except FedSage+) by at least 1\%. Additionally, the advantages of \textit{FedGrAINS} are most evident when comparing the results of FedAvg with our personalization method to those of FedPer \cite{arivazhagan2019federated}, GCFL+ \cite{GCFL}, and FedSage+ \cite{FedSage}.
Moreover, the performance of both FedSage+ and FedAvg with \textit{FedGrAINS} are very close for the Cora, CiteSeer, and PubMed datasets. However, our personalized FL framework demonstrates significantly better performance for Amazon-Computer, Amazon-Photo, and ogbn-arxiv datasets. We attribute this success to the size of the graph dataset, which resulted in a higher average degree than citation networks. 

Compared to FED-PUB \cite{FEDPUB}, FED-PUB remains superior due to its use of personalized aggregation. To further assess the impact of \textit{FedGrAINS}, we evaluate its performance when integrated with base algorithms that also employ personalized aggregation, such as FED-PUB, i.e., FED-PUB + FedGrAINS. Our results show that incorporating the importance sampler consistently and significantly enhances FED-PUB performance. Therefore, we can conclude that \textit{FedGrAINS} can be seamlessly integrated into most subgraph FL algorithms without incurring additional communication overhead or compromising privacy.

\input{sections/table-disjoint.tex}
\input{sections/table-overlap.tex}

\subsubsection*{Hyperparameter Analysis on the GFlowNet}\label{sec:gfn-hparam} To understand the operation regime of our personalization method, we vary important hyperparameters for the GFlowNet-based sampler: the
learning rate of the GFlowNet, $\beta_{GFN}$, and the reward scaling factor $\alpha$. We use the log-uniform distribution to sample the aforementioned hyperparameters with the values from the following ranges, respectively, $[1e^{-6}, 1e^{-2}]$ and $[1e^{2}, 1e^{6}]$. The setting is on the Cora dataset with the overlapping node scenario,
where we set the number of local epochs to 1 and the number of clients to 10. The results are shown in Table \ref{tab:gflownet-params}. According to Table \ref{tab:gflownet-params}, the relationship between performance and reward scaling factor $\alpha$ is highly non-linear. In addition, the learning rate $\beta_{GFN}$ is not as effective as the scaling factor $\alpha$.

\begin{table}[t!]
    \vspace{0.1in}
    \small
    \centering
    \begin{minipage}{0.49\textwidth}
        \resizebox{\textwidth}{!}{
            \renewcommand{\tabcolsep}{5.5mm}
            \renewcommand{\arraystretch}{1.2}
            \begin{tabular}{c c c  }
                \toprule 
        	    \midrule
        	    \centering
        	    \textbf{$\beta_{GFN}$} & $\alpha$ & Validation Accuracy [$\%$]  \\
        	    \midrule
                     1e-3 & 1e4 & 65.33 $\pm$ 3.11 \\
                    5e-2 & 1e4 & 68.27 $\pm$ 1.67\\
        	        1e-3 & 1e5 & \textbf{82.92} $\pm$ 1.11 \\
                    1e-2 & 1e5 & 79.42 $\pm$ 0.37\\
                    1e-3 & 1e6 & 55.27 $\pm$ 1.59 \\
                    1e-2 & 1e6 & 61.26 $\pm$ 0.97 \\
        	    \midrule
        	    \bottomrule 
            \end{tabular}
        }
    \end{minipage}
    \vspace{-0.07in}
    \caption{Sensitivity analysis on GFlowNet hyperparameters $\beta_{GFN}$ and $\alpha$. We report the hyperparameters' influence on the model performance on validation set for the Cora dataset with 10 clients in the disjoint setting.}
    \label{tab:gflownet-params}
\end{table}


%% file: sections/table-disjoint.tex
\begin{table*}[t]
\caption{\small \textbf{Results for the disjoint node scenario.} The reported results are mean and standard deviation over three different runs. The best performances in terms of accuracy are emphasized in bold.}
\label{tab:main:nonoverlap}
\small
\centering
\resizebox{\textwidth}{!}{
\begin{tabular}{lcccccccccc}
\toprule
& \multicolumn{3}{c}{\bf Cora} & \multicolumn{3}{c}{\bf CiteSeer} & \multicolumn{3}{c}{\bf PubMed}  \\
\cmidrule(l{2pt}r{2pt}){2-4} \cmidrule(l{2pt}r{2pt}){5-7} \cmidrule(l{2pt}r{2pt}){8-10} \cmidrule(l{2pt}r{2pt}){11-11}
\textbf{Methods} & \textbf{5 Clients} & \textbf{10 Clients} & \textbf{20 Clients} & \textbf{5 Clients} & \textbf{10 Clients} & \textbf{20 Clients} & \textbf{5 Clients} & \textbf{10 Clients} & \textbf{20 Clients}  \\
\midrule

Local   & 80.10 $\pm$ 0.76 & 77.43 $\pm$ 0.49 & 72.75 $\pm$ 0.89 & 70.10 $\pm$ 0.25 & 68.77 $\pm$ 0.35 & 64.51 $\pm$ 0.28 & 85.30 $\pm$ 0.24 & 84.88 $\pm$ 0.32 & 82.66 $\pm$ 0.65  \\

\midrule

FedAvg & 79.63 $\pm$ 4.37 &
72.06 $\pm$ 2.18 &
69.50 $\pm$ 3.58 &
70.24 $\pm$ 0.47 &
68.32 $\pm$ 2.59 &
65.12 $\pm$ 2.15 &
84.87 $\pm$ 0.41 &
78.92 $\pm$ 0.39 &
78.21 $\pm$ 0.25 \\

FedAvg + \textbf{\textit{FedGrAINS}} &  82.92 $\pm$ 1.11  & 80.16 $\pm$ 1.62  & 81.17 $\pm$ 1.22 & 73.09 $\pm$ 1.55  & 74.63 $\pm$ 1.35 & 70.19 $\pm$ 2.55 &85.83  $\pm$ 0.35  & 85.72 $\pm$  0.46& 84.39 $\pm$ 0.46  \\
\midrule
FedPer &  81.33 $\pm$  1.24 &
78.76 $\pm$  0.25 &
78.24 $\pm$  0.36 &
70.36 $\pm$  0.34 &
70.31 $\pm$  0.36 &
66.95 $\pm$  0.46 &
85.88 $\pm$  0.25 &
85.62 $\pm$  0.23 &
84.90 $\pm$ 0.37 \\  
FedPer + \textbf{\textit{FedGrAINS}} &  82.15 $\pm$  0.67 &
80.16 $\pm$  0.34 &
79.33 $\pm$  0.43 &
72.15 $\pm$  0.41 &
71.23 $\pm$  0.49 &
69.15 $\pm$  0.63 &
87.91 $\pm$  0.32 &
87.13 $\pm$  0.36 &
86.05 $\pm$ 0.43  \\\midrule

GCFL+ & 80.36 $\pm$  0.57 &
78.37 $\pm$  0.89 &
77.19 $\pm$  1.30 &
70.52 $\pm$  0.64 &
69.71 $\pm$  0.79 &
66.80 $\pm$  0.95 &
85.77 $\pm$  0.38 &
84.94 $\pm$  0.35 &
84.10 $\pm$  0.43  \\
GCFL+ + \textbf{\textit{FedGrAINS}} & 82.44 $\pm$ 1.15  & 79.46 $\pm$ 1.61  & 81.42 $\pm$ 1.40 & 72.79 $\pm$ 1.14 & 74.64 $\pm$ 1.48 & 70.56 $\pm$ 2.18 & 85.68 $\pm$ 0.47  & 85.37 $\pm$ 0.67 & 84.29 $\pm$ 0.51  \\
\midrule
FedSage+ & 80.09 $\pm$ 1.28 &
74.07 $\pm$ 1.46 &
72.68 $\pm$ 0.95 &
70.94 $\pm$ 0.21 &
69.03 $\pm$ 0.59 &
65.20 $\pm$ 0.73 &
86.03 $\pm$ 0.28 &
82.89 $\pm$ 0.37 &
79.71 $\pm$ 0.35 \\ 
\midrule
FED-PUB & 83.72 $\pm$ 0.18 &
81.45 $\pm$ 0.12 &
81.10 $\pm$ 0.64 &
72.40 $\pm$ 0.26 &
71.83 $\pm$ 0.61 &
66.89 $\pm$ 0.14 &
86.81$\pm$ 0.12 &
86.09$\pm$ 0.17 &
84.66 $\pm$ 0.54  \\
FED-PUB + \textbf{\textit{FedGrAINS}}   &  \textbf{84.23 $\pm$ 0.28}  & \textbf{82.68 $\pm$ 0.22}  & \textbf{81.27 $\pm$ 0.61} & \textbf{73.57 $\pm$ 0.31} & \textbf{72.48 $\pm$ 0.42} & \textbf{68.61 $\pm$ 0.19} & \textbf{88.11 $\pm$ 0.39}  & \textbf{87.91 $\pm$ 0.57}  & \textbf{87.23 $\pm$ 0.59 }  \\ 

\midrule

& \multicolumn{3}{c}{\bf Amazon-Computer} & \multicolumn{3}{c}{\bf Amazon-Photo} & \multicolumn{3}{c}{\bf ogbn-arxiv}  \\
\cmidrule(l{2pt}r{2pt}){2-4} \cmidrule(l{2pt}r{2pt}){5-7} \cmidrule(l{2pt}r{2pt}){8-10} \cmidrule(l{2pt}r{2pt}){11-11}
\textbf{Methods} & \textbf{5 Clients} & \textbf{10 Clients} & \textbf{20 Clients} & \textbf{5 Clients} & \textbf{10 Clients} & \textbf{20 Clients} & \textbf{5 Clients} & \textbf{10 Clients} & \textbf{20 Clients} \\
\midrule

Local  & 89.18 $\pm$ 0.15 &  88.25 $\pm$ 0.21 & 84.34 $\pm$ 0.28 &
91.85 $\pm$ 0.12 & 89.56 $\pm$ 0.09 & 85.83 $\pm$ 0.17 & 66.87 $\pm$ 0.09 & 66.03 $\pm$ 0.14
& 65.43 $\pm$ 0.21 \\

\midrule

FedAvg  & 88.03 $\pm$ 1.68 &
81.82 $\pm$ 1.71 &
78.19 $\pm$ 0.86 &
89.26 $\pm$ 1.80 &
85.31 $\pm$ 1.67 &
82.59 $\pm$ 1.18 &
66.24 $\pm$ 0.45 &
64.09 $\pm$ 0.83  &
62.47 $\pm$ 1.19 \\
FedAvg + \textbf{\textit{FedGrAINS}}  &  87.59 $\pm$ 0.66  & 85.21 $\pm$ 0.42  & 83.18 $\pm$ 0.11 & \textbf{92.95 $\pm$ 0.29} & \textbf{91.48 $\pm$ 0.60} & 90.60 $\pm$ 0.79 & 68.13 $\pm$ 0.58 & 66.65 $\pm$ 0.45 & 66.21 $\pm$ 0.72 \\ \midrule

FedPer  & 88.94 $\pm$ 0.25  &
88.26 $\pm$ 0.17 &
87.85 $\pm$ 0.29 &
91.30 $\pm$ 0.33 &
89.97 $\pm$ 0.27 &
88.30 $\pm$ 0.18 &
67.02 $\pm$ 0.19 &
66.02 $\pm$ 0.27 &
65.25 $\pm$ 0.31 \\ 
FedPer + \textbf{\textit{FedGrAINS}} & 89.45 $\pm$ 0.34 &
88.75 $\pm$ 0.27 &
88.33 $\pm$ 0.44 &
92.07 $\pm$ 0.38 &
90.75 $\pm$ 0.37 &
89.07 $\pm$ 0.22 &
67.68 $\pm$ 0.27 &
66.45$\pm$ 0.39 &
65.93 $\pm$ 0.42 \\ \midrule
GCFL+    & 89.07 $\pm$ 0.45
& 88.74 $\pm$ 0.49
& 87.81 $\pm$ 0.36
& 90.78 $\pm$ 0.69
& 90.22 $\pm$ 0.85
& 89.23 $\pm$ 1.07
& 66.97 $\pm$ 0.11
& 66.38 $\pm$ 0.14
& 65.30 $\pm$ 0.34 \\
GCFL+ + \textbf{\textit{FedGrAINS}}  & 89.63 $\pm$ 0.72  & 89.13 $\pm$ 0.81  & 88.62 $\pm$ 0.54 & 91.14 $\pm$ 0.47  & 90.73 $\pm$ 0.62 & 89.93 $\pm$ 0.67 & 68.14 $\pm$ 0.44 & 67.82 $\pm$ 0.45 &  67.15 $\pm$ 0.56 \\ \midrule
FedSage+ & 89.78 $\pm$ 0.71 &
84.39 $\pm$ 1.06 &
79.75 $\pm$ 0.90 &
90.89 $\pm$ 0.44 &
86.82 $\pm$ 0.78 &
83.10 $\pm$ 0.70 &
66.91 $\pm$ 0.12 &
65.30 $\pm$ 0.13 &
62.63 $\pm$ 0.24 \\ 
\midrule
FED-PUB & 90.25 $\pm$ 0.07 &
89.73 $\pm$ 0.16 &
88.20 $\pm$ 0.18 &
93.20 $\pm$ 0.15 &
92.46 $\pm$ 0.19 &
90.59 $\pm$ 0.35 &
67.62 $\pm$ 0.11 &
66.35 $\pm$ 0.16 &
63.90 $\pm$ 0.27 \\
FED-PUB + \textbf{\textit{FedGrAINS}}   &  \textbf{91.24 $\pm$ 0.23}  & \textbf{90.41 $\pm$ 0.25}  & \textbf{88.95 $\pm$ 0.15} & \textbf{94.17 $\pm$ 0.33} & \textbf{93.23 $\pm$ 0.60} & \textbf{91.42 $\pm$ 0.18} & \textbf{68.86 $\pm$ 0.43} & \textbf{68.44 $\pm$ 0.56} & \textbf{66.43 $\pm$ 0.77} \\ \midrule
\end{tabular}
}
\end{table*}




%% file: sections/table-overlap.tex
\begin{table*}[t]
\caption{\small \textbf{Results for the overlapping node scenario.} The reported results are mean and standard deviation over three different runs. The best performances in terms of accuracy are emphasized in bold.}
\label{tab:main:overlap}
\small
\centering
\resizebox{\textwidth}{!}{
\renewcommand{\arraystretch}{1.0}
\begin{tabular}{lcccccccccc}
\toprule
& \multicolumn{3}{c}{\bf Cora} & \multicolumn{3}{c}{\bf CiteSeer} & \multicolumn{3}{c}{\bf Pubmed}  \\
\cmidrule(l{2pt}r{2pt}){2-4} \cmidrule(l{2pt}r{2pt}){5-7} \cmidrule(l{2pt}r{2pt}){8-10} \cmidrule(l{2pt}r{2pt}){11-11}
\textbf{Methods} & \textbf{10 Clients} & \textbf{30 Clients} & \textbf{50 Clients} & \textbf{10 Clients} & \textbf{30 Clients} & \textbf{50 Clients} & \textbf{10 Clients} & \textbf{30 Clients} & \textbf{50 Clients} \\
\midrule

Local   & 78.14 $\pm$ 0.15 & 73.60 $\pm$ 0.18 & 69.87 $\pm$ 0.40 & 68.94 $\pm$ 0.29 & 66.13 $\pm$ 0.49
& 63.70 $\pm$ 0.92 & 84.90 $\pm$ 0.05 & 83.27 $\pm$ 0.24 & 80.88 $\pm$ 0.19\\
\midrule

FedAvg  & 78.55 $\pm$ 0.49 & 69.56 $\pm$ 0.79 & 65.19 $\pm$ 3.88 & 68.73 $\pm$ 0.46 & 65.02 $\pm$ 0.59
& 63.85 $\pm$ 1.31 & 84.66 $\pm$ 0.11 & 80.62 $\pm$ 0.46 & 80.18 $\pm$ 0.50 \\
FedAvg + \textbf{\textit{FedGrAINS}} & 79.52 $\pm$ 1.25  & 76.99 $\pm$ 1.69  & 79.45 $\pm$ 2.28 & 70.50 $\pm$ 0.93 & \textbf{69.80} $\pm$ 1.78 & 70.33 $\pm$ 2.47 & 84.59 $\pm$ 0.35& 84.32 $\pm$ 0.51 & 83.85 $\pm$ 0.69  \\
\midrule
FedPer  & 78.84 $\pm$ 0.32 &
73.46 $\pm$ 0.43 & 
72.54 $\pm$ 0.52 & 
70.42 $\pm$ 0.26 & 
65.09 $\pm$ 0.48 &
64.04 $\pm$ 0.46 & 
85.76 $\pm$ 0.14 & 
83.45 $\pm$ 0.15 &
81.90 $\pm$ 0.23 \\
FedPer + \textbf{\textit{FedGrAINS}}  & 79.73 $\pm$ 0.47 &
76.32 $\pm$ 0.63 & 
75.83 $\pm$ 0.55 & 
71.12 $\pm$ 0.37 & 
67.23 $\pm$ 0.68 &
65.85 $\pm$ 0.51 & 
86.43 $\pm$ 0.27 & 
85.15 $\pm$ 0.38 &
83.20 $\pm$ 0.47
\\ \midrule
GCFL+    & 78.60  $\pm$ 0.25 & 
73.41  $\pm$ 0.36 & 
73.13  $\pm$ 0.87 &
69.80  $\pm$ 0.34 &
65.17  $\pm$ 0.32 &
64.71  $\pm$ 0.67 & 
85.08  $\pm$ 0.21 & 
83.77  $\pm$ 0.17 & 
80.95  $\pm$ 0.22  \\
GCFL+  + \textbf{\textit{FedGrAINS}}  & 79.63 $\pm$ 1.01 & 77.01 $\pm$ 1.81 & \textbf{79.81} $\pm$ 2.82 & 70.51 $\pm$ 1.80 & 69.29 $\pm$ 1.77 & \textbf{70.45} $\pm$ 2.66 & 84.51 $\pm$ 0.34  & 84.25 $\pm$ 0.62 & 83.81 $\pm$ 0.62 \\ \midrule
FedSage+ & 79.01 $\pm$ 0.30  & 
72.20 $\pm$ 0.76 & 
66.52 $\pm$ 1.37 & 
70.09 $\pm$ 0.26 &
66.71 $\pm$ 0.18 & 
64.89 $\pm$ 0.25 &
86.07 $\pm$ 0.06 & 
83.26 $\pm$ 0.08 &
80.48 $\pm$ 0.20  \\ 
\midrule
FED-PUB & 79.65 $\pm$  0.17 & 
75.42 $\pm$  0.48 &
73.13 $\pm$  0.29 &
70.43 $\pm$  0.27 &
67.41 $\pm$  0.36 & 
65.13 $\pm$  0.40 & 
85.60 $\pm$  0.10& 
85.19 $\pm$  0.15& 
84.26 $\pm$  0.19 \\
FED-PUB + \textbf{\textit{FedGrAINS}}   &  \textbf{81.48} $\pm$ 0.23  & \textbf{77.85} $\pm$ 0.41  & 75.71 $\pm$ 0.33 & \textbf{71.95} $\pm$ 0.37 & 68.45 $\pm$ 0.40 & 67.25 $\pm$ 0.59 & \textbf{86.74} $\pm$ 0.45  &  \textbf{86.08} $\pm$ 0.57 &  \textbf{85.72} $\pm$ 0.43 \\

\midrule

& \multicolumn{3}{c}{\bf Amazon-Computer} & \multicolumn{3}{c}{\bf Amazon-Photo} & \multicolumn{3}{c}{\bf ogbn-arxiv}  \\
\cmidrule(l{2pt}r{2pt}){2-4} \cmidrule(l{2pt}r{2pt}){5-7} \cmidrule(l{2pt}r{2pt}){8-10} \cmidrule(l{2pt}r{2pt}){11-11}
\textbf{Methods} & \textbf{10 Clients} & \textbf{30 Clients} & \textbf{50 Clients} & \textbf{10 Clients} & \textbf{30 Clients} & \textbf{50 Clients} & \textbf{10 Clients} & \textbf{30 Clients} & \textbf{50 Clients} \\
\midrule

Local   & 88.50 $\pm$ 0.20 & 86.66 $\pm$ 0.00 & 87.04 $\pm$ 0.02 & 92.17 $\pm$ 0.12 & 90.16 $\pm$ 0.12 & 90.42 $\pm$ 0.15 & 62.52 $\pm$ 0.07 & 61.32 $\pm$ 0.04 & 60.04 $\pm$ 0.04 \\

\midrule

FedAvg  & 88.99 $\pm$ 0.19 & 83.37 $\pm$ 0.47 & 76.34 $\pm$ 0.12 & 92.91 $\pm$ 0.07 & 89.30 $\pm$ 0.22 & 74.19 $\pm$ 0.57 & 63.56 $\pm$ 0.02 & 59.72 $\pm$ 0.06 & 60.94 $\pm$ 0.24 \\
FedAvg + \textbf{\textit{FedGrAINS}}  & 89.25 $\pm$ 0.32  & 86.26 $\pm$ 0.62 & 85.43 $\pm$ 0.62 & 93.05 $\pm$ 0.37 & 91.05 $\pm$ 0.52 & 87.81 $\pm$ 0.88 & 65.01 $\pm$ 0.23  & 61.36 $\pm$ 0.57 & 61.04 $\pm$ 0.35   \\ \midrule
FedPer  & 89.30 $\pm$ 0.04 &
87.99 $\pm$ 0.23 &
88.22 $\pm$ 0.27 &
92.88 $\pm$ 0.24 &
91.23 $\pm$ 0.16 &
90.92 $\pm$ 0.38 &
63.97 $\pm$ 0.08 &
62.29 $\pm$ 0.04 &
61.24 $\pm$ 0.11 \\
FedPer + \textbf{\textit{FedGrAINS}}  & 89.87 $\pm$ 0.41 &
89.23 $\pm$ 0.34 &
88.41 $\pm$ 0.33 &
93.27 $\pm$ 0.31 &
91.71 $\pm$ 0.23 &
91.02 $\pm$ 0.41 &
64.04 $\pm$ 0.11 &
63.03 $\pm$ 0.21 &
61.93 $\pm$ 0.27  
\\ \midrule
GCFL+    & 89.01 $\pm$ 0.22 &
87.24 $\pm$ 0.09 &
87.02 $\pm$ 0.22 &
92.45 $\pm$ 0.10 & 
90.58 $\pm$ 0.11 &
90.54 $\pm$ 0.08 &
63.24 $\pm$ 0.02 &
61.66 $\pm$ 0.10 &
60.32 $\pm$ 0.01 \\
GCFL+  + \textbf{\textit{FedGrAINS}}  & 89.82 $\pm$ 0.66  & 89.13 $\pm$ 0.45 & 88.41 $\pm$ 0.45 & 92.75 $\pm$ 0.34 & 91.07 $\pm$ 0.28 & 90.69 $\pm$ 0.46  & 64.22 $\pm$ 0.13 & 63.35 $\pm$ 0.27 & 62.29 $\pm$ 0.37  \\ \midrule
FedSage+ & 89.24 $\pm$ 0.15 &
81.33 $\pm$ 1.20 &
76.72 $\pm$ 0.39 &
92.76 $\pm$ 0.05 &
88.69 $\pm$ 0.99 &
72.41 $\pm$ 1.36 &
63.24 $\pm$ 0.02 &
59.90 $\pm$ 0.12 &
60.95 $\pm$ 0.09 \\  \midrule
FED-PUB     & 89.98 $\pm$ 0.08 & 
89.15 $\pm$ 0.06 &
88.76 $\pm$ 0.14 & 
93.22 $\pm$ 0.07 & 
92.01 $\pm$ 0.07 & 
\textbf{91.71} $\pm$ 0.11 & 
64.18 $\pm$ 0.04 & 
63.34 $\pm$ 0.12 & 
62.55 $\pm$ 0.12 \\
FED-PUB + \textbf{\textit{FedGrAINS}}  &  \textbf{90.41}$\pm$ 0.23  & \textbf{90.05} $\pm$ 0.28  & \textbf{89.73} $\pm$ 0.13 & \textbf{93.36} $\pm$ 0.21 & \textbf{92.85} $\pm$ 0.60 & 91.45 $\pm$ 0.29 & \textbf{67.38} $\pm$ 0.25  & \textbf{65.13} $\pm$ 0.33  & \textbf{64.17} $\pm$ 0.17   \\
\bottomrule

\end{tabular}
}
\end{table*}


%% file: sections/related.tex
\section{Related Works}
\label{related_work} 

\subsubsection*{GNNs and Random dropping GNNs:} GNNs are the de-facto tools to learn the representations of nodes, edges, and entire graphs~\cite{GNN, GNN/1, GNN/2, edge, gmt}. Most existing GNNs fall under the message-passing neural network framework~\cite{MPNN}, which iteratively represents a node representation by aggregating features from its neighboring nodes and itself~\cite{GCN, spectralgraph, GraphSAGE}. However, the current paradigm for GNNs is restrictive regarding the development of deeper architectures due to the issues of over-smoothing and overfitting \cite{li2018deeper, GCN}. To remedy this, dropout and its variants are utilized by
randomly removing the node features, edges, or messages during training \cite{dropout, Rong2020DropEdge,fang2023dropmessage}. However, these regularizers are independent of the graph topology, limiting their efficacy under data and structural heterogeneity. Moreover, state-of-the-art GNNs on node classification and link prediction tasks are tailored for a single graph, limiting their applicability in real-world systems with locally distributed graphs that are not shared across participants, which gives rise to the use of FL for GNNs \cite{fedgraph,he2021fedgraphnn}.

\subsection*{Federated Learning (FL):} Federated Learning allows multiple users to learn a collaborative model under the guidance of a centralized server without sharing raw data \cite{McMahan2017CommunicationEfficientLO, FLsurvey}. One of the primary bottlenecks in this learning paradigm is data heterogeneity across different entities. To address data heterogeneity, proximal term regularization has been proposed to prevent clients' local models from diverging excessively from their local training data \cite{li2020federated}. For clients with extreme levels of heterogeneity, learning a single global model may not be suitable. Personalizing client models relaxes the constraint of developing one global model, allowing for the optimization of average empirical risk across clients through various methods, such as distillation \cite{fedgkt}, meta-learning \cite{fallah2020personalized}, and knowledge separation \cite{arivazhagan2019federated}. However, graph-structured data presents additional challenges due to the connections between instances, leading to missing edges and community structures within private subgraphs, unlike more common modalities like images and text \cite{he2021fedgraphnn, FedSage}.

\subsection*{FL over Graphs:} 
Recent studies primarily categorize the use of FL frameworks for training GNNs into subgraph-level and graph-level federated learning methods~\cite{he2021fedgraphnn, graphflsystem}. Graph-level methods operate under the condition that different clients have their own set of graph datasets that are not part of a bigger graph (e.g., molecular graphs), and recent works~\cite{GCFL, SpreadGNN, graphfl} focus on the label heterogeneity among clients' non-IID graphs. On the other hand, the subgraph-level FL poses a unique graph-structural challenge: the existence of missing yet probable links between subgraphs, as each subgraph is a part of the larger global graph. To deal with the aforementioned problem, existing methods~\cite{FedGNN, FedSage, FedGCN} augment the nodes by requesting the node information in the other subgraphs and then connecting the existing nodes with the augmented ones. However, requesting node information from other clients increases communication overhead and the risk of privacy leakage. Recently, \cite{FEDPUB} introduced a novel problem of heterogeneity among subgraph communities \cite{Community1, Community} across clients' subgraphs and proposed a personalized subgraph FL framework that employs personalized aggregation and sparse masking. Unlike other methods, this work focuses on adjusting node information without requiring additional data. We employ a local generative model to estimate the importance of nodes to minimize classification loss. Furthermore, our approach does not incur any extra communication overhead and serves as an off-the-shelf personalization method on top of any existing personalized (sub-)graph FL algorithms.

%% file: sections/conclusion.tex
\section{Conclusion \& Future Work}
    This work addresses the challenges of node-degree heterogeneity and local overfitting in subgraph FL by proposing \textit{FedGrAINS}, an adaptive neighborhood sampling-based regularization method. For each client, \textit{FedGrAINS} employs a GFlowNet model to assess the node-importance to maximize the node classification performance. On the server side, these node importance distributions are aggregated in a personalized manner at every round, enabling adaptive generalization. By leveraging a generative and data-driven sampling approach, \textit{FedGrAINS} ensures consistent performance, effectively handling missing links and subgraph heterogeneity in an intuitive yet elegant manner. Our experimental results demonstrate that incorporating \textit{FedGrAINS} as a regularization method significantly improves the performance and generalization of the FL subgraph techniques. 

We plan to extend our personalization framework to other federated graph learning settings \cite{SpreadGNN,FedGCN,fedgraph}. As long as we have access to a tractable reward \cite{bengio2021flow}, we can leverage the proposed \textit{FedGrAINS}. We assumed that $Z(s_0)$ is constant to reduce the workload at the edge. We defer the investigation of how the estimation of $Z(s_0)$ might enhance personalization to future work \cite{gaflow}. Finally, we would like to obtain theoretical insights into our framework using the tools presented in \cite{hasanzadeh2020bayesian}, evaluating how effectively GFlowNet alleviates the oversmoothing and over-fitting tendencies of client GNNs while allowing uncertainty quantification at the edge at no additional cost.

\section{Acknowledgements}

We thank anonymous reviewers for their constructive feedback. Emir Ceyani is grateful for the continuous support of coauthors, his family, and housemates and for Saurav Prakash's discussions on the use of GFlowNets.

%% file: sections/appendix.tex
\clearpage



\clearpage
\section{Algorithm Outline}
\label{appendix:fedgrains-algo}
In this section, we present client-side and server-side algorithms of our personalized subgraph FL framework \textit{FedGrAINS}  shown in Algorithm~\ref{algo:fedgrains_client} and Algorithm~\ref{algo:fedgrains_server}.

\begin{figure}[h!]
    \vspace{-0.2in}
    \begin{algorithm}[H]
        \small
    	\caption{\textbf{FedGrAINS} Client Algorithm}
    	    \begin{algorithmic}[1]
    	        \label{algo:fedgrains_client}

                \Require Graph $\mathcal{G}$, node features $\bm{X}$, node labels $\bm{Y}$, batch size $B$, sampling budget $k$, $GNN:$ GNN for classification, $GFN:$ GNN for GFlowNet ,  $\eta_{GNN}$: Learning rate for the GNN classifier, $\eta_{GFN}$: Learning rate for the GFlowNet,  $E$: the number of local epochs,  $G_i$: Gumbel noise for the $i^{th}$ node, $\phi_{GNN}^i$: GNN parameters for client $i$, $\phi_{GFN}^i$: GFlowNet parameters for client $i$.

                \State   \textbf{Function} RunClient($\bar{\phi}_{GNN}^i$)
                \State $ \phi_{GNN}^i \leftarrow \bar{\phi}_{GNN}^i$
                    \For{each local epoch $e$ from $1$ to $E$}
                    \For {each batch $\mathcal{V}^b$}
                    \State{$\mathcal{S}^0=\mathcal{V}^b$}
    \State{Build adjacency matrix $A^0$ from $\mathcal{S}^0$}
    \For {layer $l=1$ to $L$}
        \State {$\mathcal{V}_n^l \gets \mathcal{N}(\mathcal{S}^{l-1})$ \Comment{Get all $n$ neighbors of $\mathcal{S}^{l-1}$} }
        \State {$p_1, ..., p_{n} \gets GFN(A_0, ..., A_{l-1})$ \Comment{Compute probabilities of inclusion} }
        \State {$G_1, ..., G_{n}\sim \text{Gumbel}(0, 1)$ \Comment{Sample Gumbel noise}}
        \State {$\mathcal{V}_k^l\gets \mathsf{top}(k, \log p_1 + G_1, ..., \log p_n + G_n)$ \Comment{Get $k$ best nodes (Eq. \ref{eqn:gumbel})}}
        \State{$\mathcal{S}^{\ell}=\mathcal{S}^{\ell-1} \cup \mathcal{V}_k^l$ \Comment{Add new nodes}}
        \State{Build adjacency matrix $A^l$ from $K^l$}
    \EndFor
    \State {Pass all $\hat{A}_l$ to $GNN$ and obtain classification loss $\mathcal{L}_{GNN}$ given $\{(\bm{X}_j,\bm{Y}_j)\}_{j=1}^B$}
    \State {$R(A_0,...,A_L) \leftarrow \exp (-\alpha \cdot \mathcal{L}_{GNN})$ \Comment{Calculate reward}}
    \State Compute \textit{FedGrAINS} loss $\mathcal{L}_{FedGrAINS}$ from Eq. \ref{fedgrains}
    \State $\phi_{GNN}^i \leftarrow \bar{\phi}_{GNN}^i-\eta_{GNN}\nabla\mathcal{L}_{FedGrAINS}(\mathcal{G}; \bar{\phi}_{GNN}^i)$ \Comment{Update GNN parameters}
    \State $\phi_{GNN}^i \leftarrow \bar{\phi}_{GFN}^i-\eta_{GFN}\nabla\mathcal{L}_{FedGrAINS}(\mathcal{G}; \bar{\phi}_{GNN}^i)$ \Comment{Update GFN parameters}
\EndFor

                    \EndFor
                    \State \textbf{return} $\phi_{GNN}^i$
        	\end{algorithmic}
    	\end{algorithm}

\end{figure}

\begin{figure*}[h!]
    \vspace{-0.2in}
        \begin{algorithm}[H]
            \small
            \caption{\textbf{FedGrAINS} Server Algorithm}
        	\begin{algorithmic}[1]
        	    \label{algo:fedgrains_server}
        	\Require $R$: the number of rounds, $E$: the number of epochs, $M$: the number of clients,
                \State   \textbf{Function} RunServer()
                    \State initialize $\bar{\phi}_{GNN}^{(1)} $
                    \For{each round $r=1,2,\dots, R$}
                        \For{$\forall i~\textbf{\mbox{in parallel}}$}
                            \If{$r = 1$} 
                                \State $\phi_{GNN}^{(r+1),i} \leftarrow \text{RunClient} (\bar{\phi}_{GNN}^{(r)})$
                            \Else
                                
                                \State $\bar{\phi}_{GNN}^{(r)} \leftarrow \text{Aggregate}(\{\bar{\phi}_{GNN}^{(r),j}\}_{j=1}^{M}$)
                                \State $\phi_{GNN}^{(r+1),i} \leftarrow \text{RunClient} (\bar{\phi}_{GNN}^{(r)})$
                            \EndIf 
                    	\EndFor
                    \EndFor 
             
        	\end{algorithmic}
    	\end{algorithm}    
\end{figure*}

\section{Experimental Setups}
\label{appendix:setup}

In this section, we first describe six benchmark datasets utilized in our study, including their preprocessing setups for federated learning and relevant statistics in Subsection~\ref{appendix:sub:data}. Next, we provide detailed explanations of the baselines and our proposed \textit{FedGrAINS} in Subsection~\ref{appendix:sub:model}.

\begin{table*}[t]
\caption{\small \textbf{Dataset statistics.} We report the number of nodes, edges, classes, clustering coefficient, and heterogeneity for the original graph and its split subgraphs on overlapping and disjoint node scenarios. Ori denotes the original graph, and Cli denotes the number of clients.}
\label{tab:appendix:data}
\small
\centering
\resizebox{\textwidth}{!}{
\renewcommand{\arraystretch}{0.9}
\begin{tabular}{lcccccccccccc}
\toprule
\multicolumn{13}{l}{\bf \emph{Overlapping node scenario}} \\
\midrule
& \multicolumn{4}{c}{\bf Cora} & \multicolumn{4}{c}{\bf CiteSeer} & \multicolumn{4}{c}{\bf Pubmed}  \\
\cmidrule(l{2pt}r{2pt}){2-5} \cmidrule(l{2pt}r{2pt}){6-9} \cmidrule(l{2pt}r{2pt}){10-13}
& \textbf{Ori} & \textbf{10 Cli} & \textbf{30 Cli} & \textbf{50 Cli} & \textbf{Ori} & \textbf{10 Cli} & \textbf{30 Cli} & \textbf{50 Cli} & \textbf{Ori} & \textbf{10 Cli} & \textbf{30 Cli} & \textbf{50 Cli} \\
\midrule

\# Classes & \multicolumn{4}{c}{7} & \multicolumn{4}{c}{6} & \multicolumn{4}{c}{3} \\
\# Nodes   & 2,485 & 621 & 207 & 124 & 2,120 & 530 & 177 & 106 & 19,717 & 4,929 & 1,643 & 986 \\
\# Edges   & 10,138 & 1,249 & 379 & 215 & 7,358 & 889 & 293 & 170  & 88,648 & 10,675 & 3,374 & 1,903 \\
Clustering Coefficient & 0.238 & 0.133 & 0.129 & 0.125 & 0.170 & 0.088 & 0.087 & 0.096 & 0.060 & 0.035 & 0.034 & 0.035 \\
Heterogeneity & N/A & 0.297 & 0.567 & 0.613 & N/A & 0.278 & 0.494 & 0.547 & N/A & 0.210 & 0.383 & 0.394 \\

\midrule

& \multicolumn{4}{c}{\bf ogbn-arxiv} & \multicolumn{4}{c}{\bf Amazon-Computer} & \multicolumn{4}{c}{\bf Amazon-Photo} \\
\cmidrule(l{2pt}r{2pt}){2-5} \cmidrule(l{2pt}r{2pt}){6-9} \cmidrule(l{2pt}r{2pt}){10-13}
& \textbf{Ori} & \textbf{10 Cli} & \textbf{30 Cli} & \textbf{50 Cli} & \textbf{Ori} & \textbf{10 Cli} & \textbf{30 Cli} & \textbf{50 Cli} & \textbf{Ori} & \textbf{10 Cli} & \textbf{30 Cli} & \textbf{50 Cli} \\
\midrule

\# Classes & \multicolumn{4}{c}{40} & \multicolumn{4}{c}{10} & \multicolumn{4}{c}{8} \\
\# Nodes   & 169,343 & 42,336 & 14,112 & 8,467 & 13,381 & 3,345 & 1,115 & 669 & 7,487 & 1,872 & 624 & 374 \\
\# Edges   & 2,315,598 & 282,083 & 83,770 & 44,712 & 491,556 & 59,236 & 16,684 & 8,969 & 238,086 & 29,223 & 8,735 & 4,840 \\
Clustering Coefficient & 0.226 & 0.177 & 0.185 & 0.191 & 0.351 & 0.337 & 0.348 & 0.359 & 0.410 & 0.380 & 0.391 & 0.410  \\
Heterogeneity & N/A & 0.315 & 0.606 & 0.615 & N/A & 0.327 & 0.577 & 0.614 & N/A & 0.306 & 0.696 & 0.684 \\

\midrule
\midrule

\multicolumn{13}{l}{\bf \emph{Non-overlapping node scenario}} \\
\midrule
& \multicolumn{4}{c}{\bf Cora} & \multicolumn{4}{c}{\bf CiteSeer} & \multicolumn{4}{c}{\bf Pubmed}  \\
\cmidrule(l{2pt}r{2pt}){2-5} \cmidrule(l{2pt}r{2pt}){6-9} \cmidrule(l{2pt}r{2pt}){10-13}
& \textbf{Ori} & \textbf{5 Cli} & \textbf{10 Cli} & \textbf{20 Cli} & \textbf{Ori} & \textbf{5 Cli} & \textbf{10 Cli} & \textbf{20 Cli} & \textbf{Ori} & \textbf{5 Cli} & \textbf{10 Cli} & \textbf{20 Cli} \\
\midrule

\# Classes &  \multicolumn{4}{c}{7} & \multicolumn{4}{c}{6} & \multicolumn{4}{c}{3} \\
\# Nodes   & 2,485 & 497 & 249 & 124 & 2,120 & 424 & 212 & 106 & 19,717 & 3,943 & 1,972 & 986 \\
\# Edges   & 10,138 & 1,866 & 891 & 422 & 7,358 & 1,410 & 675 & 326 & 88,648 & 16,374 & 7,671 & 3,607 \\
Clustering Coefficient & 0.238 & 0.250 & 0.259 & 0.263 & 0.170 & 0.175 & 0.178 & 0.180 & 0.060 & 0.063 & 0.066 & 0.067 \\
Heterogeneity & N/A & 0.590 & 0.606 & 0.665 & N/A & 0.517 & 0.541 & 0.568 & N/A & 0.362 & 0.392 & 0.424 \\

\midrule

& \multicolumn{4}{c}{\bf ogbn-arxiv} & \multicolumn{4}{c}{\bf Amazon-Computer} & \multicolumn{4}{c}{\bf Amazon-Photo} \\
\cmidrule(l{2pt}r{2pt}){2-5} \cmidrule(l{2pt}r{2pt}){6-9} \cmidrule(l{2pt}r{2pt}){10-13}
& \textbf{Ori} & \textbf{5 Cli} & \textbf{10 Cli} & \textbf{20 Cli} & \textbf{Ori} & \textbf{5 Cli} & \textbf{10 Cli} & \textbf{20 Cli} & \textbf{Ori} & \textbf{5 Cli} & \textbf{10 Cli} & \textbf{20 Cli} \\
\midrule

\# Classes &  \multicolumn{4}{c}{40} & \multicolumn{4}{c}{10} & \multicolumn{4}{c}{8} \\
\# Nodes   & 169,343 & 33,869 & 16,934 & 8,467 & 13,381 & 2,676 & 1,338 & 669 & 7,487 & 1,497 & 749 & 374 \\
\# Edges   & 2,315,598 & 410,948 & 182,226 & 86,755 & 491,556 & 84,480 & 36,136 & 15,632 & 238,086 & 43,138 & 19,322 & 8,547  \\
Clustering Coefficient & 0.226 & 0.247 & 0.259 & 0.269 & 0.351 & 0.385 & 0.398 & 0.418 & 0.410 & 0.437 & 0.457 & 0.477 \\
Heterogeneity & N/A & 0.593 & 0.615 & 0.637 & N/A & 0.604 & 0.612 & 0.647 & N/A & 0.684 & 0.681 & 0.751 \\

\bottomrule

\end{tabular}
}
\end{table*}

\subsection{Datasets}
\label{appendix:sub:data}

We report statistics of six different benchmark datasets~\cite{planetoid, ogb, amazon, amnazonsubset}, such as Cora, CiteSeer, Pubmed, and ogbn-arxiv for citation graphs; Computer and Photo for amazon product graphs, which we use in our experiments, for both the overlapping and non-overlapping node scenarios in Table~\ref{tab:appendix:data}. Specifically, in Table~\ref{tab:appendix:data}, we report the number of nodes, edges, classes, and clustering coefficient for each subgraph, but also the heterogeneity between the subgraphs. Note that, to measure the clustering coefficient, which indicates how much nodes are clustered together, for the individual subgraph, we first calculate the clustering coefficient for all nodes, and then average them. On the other hand, to measure the heterogeneity, which indicates how disjointed subgraphs are dissimilar, we calculate the median Jenson-Shannon divergence of label distributions between all pairs of subgraphs.

\subsubsection{Missing Links}
\label{appendix:sub:data:links}
In Table 5 and 6, we show the total number of missing links in the non-overlapping and overlapping
settings with different numbers of clients. We can observe that there are more missing links with
the increased number of clients, which leads to more severe information loss in subgraph federated
learning. We also note that the overlapping setting has more missing links than the non-overlapping
setting with the same number of clients, e.g., 10 clients. This is mainly due to the node sampling
procedure in the overlapping setting. Some links in the partitioned graph are not sampled and included
in any client.
\begin{table}
\centering
\begin{tabular}{c|ccc}
\hline Dataset  & 5 clients  & 10 clients  & 20  clients  \\
\hline  Cora & 403 & 615 & 853 \\
 CiteSeer  & 105 & 304 & 424 \\
PubMed  & 3,388 & 5,969 & 8,254 \\
 Amazon-Computer  & 34,578 & 65,098 & 89,456 \\
Amazon-Photo  & 28,928 & 22,434 & 33,572 \\
ogbn-arxiv  & 130,428 & 246,669 & 290,249 \\
\hline
\end{tabular}
\caption{ Total number of missing links with different numbers of clients (non-overlapping).}
\end{table}

\begin{table}
\centering
\begin{tabular}{c|ccc}
\hline Dataset & 10 clients & 30 clients & 50 clients \\
\hline Cora & 1,391 & 1,567 & 1,733 \\
CiteSeer & 922 & 1,043 & 1,160 \\
PubMed & 11,890 & 13,630 & 15,060 \\
Amazon-Computer & 66,562 & 95,530 & 107,740 \\
Amazon-Photo & 11,197 & 37,207 & 45,219 \\
ogbn-arxiv & 291,656 & 392,895 & 457,954 \\
\hline
\end{tabular}
\caption{ Total number of missing links with respect to the number of clients (disjoint)}
\end{table}
\subsubsection{Node Degree Heterogeneity}
\label{appendix:sub:data:degree-het}
Tables 7 and 8 show the node-degree heterogeneity with different numbers of clients in non-overlapping and
overlapping settings. To measure the node-degree heterogeneity, we calculate the average Hellinger distance of
degree distributions over all client pairs except with themselves. We can observe that a larger number of clients
leads to more severe node-degree heterogeneity and the non-overlapping setting is more heterogeneous than
the overlapping setting. This results in poor performance with GNNs. 

\begin{table}
\centering
\begin{tabular}{c|ccc}
\hline Dataset  & 5 clients  & 10 clients  & 20  clients  \\
\hline  Cora & 0.4292 & 0.6508 & 0.7198 \\
 CiteSeer  & 0.4864
 & 0.6507
& 0.7076\\
PubMed  & 0.1333 &
0.1515
 & 0.2279 \\
 Amazon-Computer  & 0.265 &
0.2526 &
0.2855\\
Amazon-Photo  & 0.234 &
0.2438 &
0.3079\\
ogbn-arxiv  & 0.1595 &
0.1762 &
0.1976 \\
\hline
\end{tabular}
\caption{ Average pairwise Hellinger distance between degree distributions with different numbers of clients (non-overlapping).}
\end{table}

\begin{table}
\centering
\begin{tabular}{c|ccc}
\hline Dataset & 10 clients & 30 clients & 50 clients \\
\hline Cora & 0.4265 &
0.7237 &
0.7492 \\
CiteSeer & 0.6739 &
0.7314 &
0.75 \\
PubMed & 0.1428 &
0.3875 &
0.4352\\
Amazon-Computer & 0.2623 &
0.258 &
0.2682
\\
Amazon-Photo & 0.3009 &
0.2356 &
0.2942 \\
ogbn-arxiv & 
0.1442 &
0.2071 &
0.2017\\
\hline
\end{tabular}
\caption{Average pairwise Hellinger distance between degree distribtutions with different numbers of clients (disjoint).}
\end{table}

\subsection{Baselines and Our Model}
\label{appendix:sub:model}

\begin{enumerate}

     \item \textbf{Local}: This method is the non-FL baseline, which only locally trains the model for each client without weight sharing.

    \item \textbf{FedAvg}: This method~\cite{McMahan2017CommunicationEfficientLO} is the FL baseline, where each client locally updates a model and sends it to a server, while the server aggregates the locally updated models with respect to their numbers of training samples and transmits the aggregated model back to the clients.
    
    \item \textbf{FedPer}: This method~\cite{arivazhagan2019federated} is the personalized FL baseline, which shares only the base layers, while keeping the personalized classification layers in the local side.
    
    \item \textbf{FedSage+}: This method~\cite{FedSage} is the subgraph FL baseline, which expands local subgraphs by generating additional nodes with the local graph generator. To train the graph generator, each client first receives node representations from other clients, and then calculates the gradient of distances between the local node features and the other client's node representations. Then, the gradient is sent back to other clients, which is then used to train the graph generator.
    
    \item \textbf{GCFL}: This method~\cite{GCFL} is the graph FL baseline, which targets completely disjoint graphs (e.g., molecular graphs) as in image tasks. In particular, it uses the bi-partitioning scheme, which divides a set of clients into two disjoint client groups based on their gradient similarities. Then, the model weights are only shared between grouped clients having similar gradients, after partitioning. Note that this bi-partitioning mechanism is similar to the mechanism proposed in clustered-FL~\cite{sattler2020clustered} for image classification, and we adopt this for our subgraph FL.
   
    \item \textbf{FED-PUB}: This method \cite{FEDPUB} performs personalized aggregation based on subgraph similarities and models' functional embeddings for discovering community structures, but also adaptively masks the received weights from the server to filter irrelevant weights from heterogeneous communities.

    \item \textbf{\textit{FedGrAINS}}: Our plug-and-play  personalized subgraph federated learning model, which employs a personalized GFlowNet for each client. For each node in the client's graph, the GFlowNet estimates the important nodes over the $k-$hop neighborhood and samples using Gumbel-max trick\cite{huijben2022review-gumbel,kool2019stochastic} so that the classification performance optimizes. Then. back-propagating through our novel loss, we update the classifier and the node-importance sampler simultaneously. This algorithm only incurs extra local computation while introducing no additional communication cost and privacy breach.

\end{enumerate}